\documentclass[11pt]{article}

\usepackage[margin=1in]{geometry}
\usepackage[T1]{fontenc}
\usepackage{lmodern}
\usepackage{microtype}
\usepackage{amsmath,amssymb}
\usepackage{booktabs}
\usepackage{array}
\usepackage{graphicx}
\usepackage{xcolor}
\usepackage{enumitem}
\usepackage{caption}
\usepackage{siunitx}
\usepackage{pgfplots}
\pgfplotsset{compat=1.17}
\usepackage[colorlinks=true,linkcolor=blue!55!black,citecolor=blue!55!black,urlcolor=blue!55!black]{hyperref}

\newcommand{\tok}[1]{\num{#1}}
\newcommand{\paritok}{\textsc{Paritok}}

\title{\textbf{An Empirical Cost Attribution of Context-Compression\\ Gateways in Multi-Turn Coding Agents}\\[4pt]
\large Why Compression Rate Is Not the End-to-End Saving}

\author{Luzhuo Chen \qquad Jiayu Shi\\[2pt]
\small Paritok\\
\small \texttt{paritok9@gmail.com}}

\date{\today}

\begin{document}
\maketitle

\begin{abstract}
Context compression is widely proposed as a way to cut the token bill of LLM-based coding agents, and public benchmarks report that aggressive compression preserves task-solving quality. Yet these two facts do not imply the third one everybody assumes: that compressing file reads actually saves money in a real multi-turn agent. We instrument a production compression gateway (\paritok) sitting between coding agents (Claude Code, Codex) and frontier LLMs (Claude Sonnet, GPT-5), and decompose the token bill of real agent sessions into three independent levers: (i) \emph{tool-schema filtering}, (ii) \emph{content compression} of file reads and tool output, and (iii) \emph{history summarization}. Measuring each lever in isolation across controlled A/B runs, we find that the three levers save money at fundamentally different rates. Tool-schema filtering saves a \emph{fixed} block of tokens every turn (linear in turn count~$N$) and is the only lever that is unambiguously and reproducibly positive; on a typical turn it removes $\sim$21K--57K tokens. Content compression, by contrast, saves only $\sim$2\% of the (cache-priced) prefix per turn---but because compressed reads accumulate in history and are re-sent on every later turn, its cumulative saving grows \emph{quadratically}, $\approx 3{,}350\,N^2$ tokens (measured), overtaking the fixed tool-filter saving within the first $\sim$6 turns before the context window caps it. Crucially, a non-destructive gateway lets the agent pull the original bytes back on demand, and whenever it does the recall simply re-sends the bytes just compressed away, so each recall costs a fixed, bounded amount---the length of the one compressed segment it re-sends, never a multiplicative blowup. The more the agent recalls, the more of the accumulated content saving is spent back, one compressed segment at a time. Finally, we show that a strong single-shot compression benchmark---the 86.5\% SWE-bench quality retained at a 25.7\% compression rate achieved by the compression \emph{model} the gateway deploys (Paritok-4B, reported separately)---is \emph{orthogonal} to multi-turn agent cost and must not be cited as a cost-saving argument for the gateway. We distill these into an actionable recipe for where token-saving effort actually pays off.
\end{abstract}

\section{Introduction}

The dominant cost of an autonomous coding agent is not the model's output---it is the \emph{input it re-sends on every turn}. A modern agent (Claude Code, Cursor, Codex, OpenHands) builds each request from three re-sent parts: a block of tool/function JSON schemas, an ever-growing message history, and large tool-result / file-read blocks. Because agents solve tasks over many turns of grep, read, edit, and test~\cite{react}, this prefix is transmitted dozens of times per task.

A natural response is \emph{context compression}: shrink the file reads and tool outputs with a small model before they reach the expensive frontier LLM~\cite{llmlingua,llmlingua2,selectivecontext}. This is attractive and increasingly popular, and a well-trained compressor can shrink each segment to a quarter of its size while a single-shot benchmark~\cite{swebench} shows the downstream model still solves the task. It is tempting to conclude that such compression proportionally cuts the bill.

\textbf{It does not.} In this paper we report a controlled empirical study of \emph{where the tokens actually go} in real multi-turn agent sessions, and of which interventions actually move the bill. Our vehicle is \paritok, an open-source non-destructive compression gateway that intercepts every request, and whose three levers---tool-schema filtering, content compression, and history summarization---can be independently toggled. This lets us do something benchmark papers cannot: attribute dollars to levers, in situ, across real agent trajectories.

Our contributions are:

\begin{enumerate}[leftmargin=1.4em,itemsep=2pt]
  \item A \textbf{cost model} for multi-turn agents (\S\ref{sec:model}) that makes explicit why turn count, not compression rate, is the first-order cost variable, and why cache pricing (a $10\times$ discount on re-sent prefix) changes which savings matter.
  \item An \textbf{isolated measurement of three levers} (\S\ref{sec:levers}) on real Claude Code and Codex sessions, showing tool filtering is the dominant and only reliably-positive lever, while per-turn content compression is a $\sim$2\% marginal effect drowned by turn-count variance.
  \item \textbf{Scaling laws} (\S\ref{sec:scaling}): tool filtering saves linearly in $N$, content compression \emph{quadratically}, with a measured crossover near $N\!\approx\!6$ and a hard cap from the context window. From a 5-turn cumulative experiment, the measured content saving fits $\approx 3{,}350\,N^2$ while the tool filter is a fixed $\sim$21K/turn.
  \item The \textbf{recall trap} (\S\ref{sec:expand}): a non-destructive gateway's biggest virtue---exact recovery---is also what eats into content-compression savings, one recall at a time: each recall re-sends the original it had just compressed, so the compressed copy is wasted---a fixed cost of one compressed segment per recall, equal to the length of that compressed text (bounded and deterministic, never a multiplicative blowup). The content lever's net is the accumulated saving minus that per-recall cost, so its sign is set by recall frequency and session length alone.
  \item A \textbf{benchmark caveat} (\S\ref{sec:swebench}): single-shot compression quality (SWE-bench Lite) is orthogonal to multi-turn cost; conflating them is the field's most common error.
  \item \textbf{Methodology} (\S\ref{sec:method}) for reproducible agent-cost measurement: distributions over per-turn attribution, provider-reported usage over self-estimates, cache-tier accounting.
\end{enumerate}

We deliberately report the sometimes deflationary picture: a raw \emph{segment} compression rate is not the end-to-end saving, which we measure at $\sim$25\% on a single turn and $\sim$39\% by turn five on a long session, reduced by a fixed cost of one compressed segment on each of the minority of turns where the agent recalls the original. This is the point: it tells a practitioner exactly where to spend effort.

\section{Setup and Cost Model}
\label{sec:model}

\subsection{Experimental harness}

\paritok{} is a proxy that speaks the Anthropic Messages, OpenAI Chat, and OpenAI Responses protocols, so a coding agent points its \texttt{BASE\_URL} at the proxy and is otherwise unchanged. On each request the gateway may (a) filter the tool-schema block down to the semantically relevant tools via a local CPU embedding model~\cite{bge}, stubbing the rest behind a recall tool; (b) compress each file read / tool result with a 4B code-native compression model (a LoRA~\cite{lora} adapter over Qwen3-4B~\cite{qwen3}) to $\sim$26\% of its size, tagging it \texttt{[REF:id]}; and (c) summarize stale history once the window fills. Every compression is \emph{non-destructive}: the agent can call a recall tool (\texttt{read\_original}, \texttt{gateway\_search\_tools}) to retrieve exact bytes on demand.

Each lever can be independently disabled, which is what makes attribution possible. To isolate content compression we set a token floor \texttt{min\_tokens} to $512$ (``compress'') versus $49999$ (``no-op''), holding tool filtering fixed; to isolate tool filtering we toggle it while holding the content path fixed; to remove the confound of MCP tool-block jitter (\S\ref{sec:method}) we run with \texttt{--strict-mcp-config}.

We run two agent families---Claude Code (upstream Claude Sonnet) and Codex (upstream GPT-5)---on real repositories (\texttt{textual}, \texttt{jinja2}, \texttt{werkzeug}, \texttt{rich}, \texttt{httpx}, \texttt{tox}). Backends are a local Ollama deployment of the 4B model and a GPU server; we verified both compress identically (segment ratio $\approx 0.44$ on line-numbered reads). All dollar figures use \emph{provider-reported} \texttt{usage}, priced per tier.

\subsection{The bill is a sum over turns, and cache pricing dominates}

Let a session run for $N$ turns. On turn $t$ the agent sends a prefix $P_t$ (system + tools + accumulated history + reads) and receives output $O_t$. Frontier providers price re-sent prefix at a \emph{cache-read} tier roughly $10\times$ cheaper than fresh input. The bill is
\begin{equation}
\text{Cost} \;=\; \sum_{t=1}^{N}\Big( c_{\text{cr}}\,P^{\text{cached}}_t + c_{\text{in}}\,P^{\text{new}}_t + c_{\text{out}}\,O_t \Big),
\label{eq:bill}
\end{equation}
with (Sonnet) $c_{\text{in}}=\$3$, $c_{\text{cr}}=\$0.30$, $c_{\text{out}}=\$15$ per million tokens. Two consequences drive the entire paper:

\begin{itemize}[leftmargin=1.4em,itemsep=2pt]
  \item \textbf{Turn count is the first-order variable.} The prefix $P_t$ grows with the conversation and is paid every turn, so $\text{Cost}\approx \sum_t P_t$ is dominated by \emph{how many turns} the agent takes---an emergent property of agent behavior (how much it reads, edits, explores), largely orthogonal to compression. Empirically, the same task on the same config swings from 7 to 18 turns; this variance dwarfs a 2\% per-turn effect.
  \item \textbf{Not all saved tokens cost the same.} A token removed from the re-sent prefix is a \emph{cache-read} token worth $c_{\text{cr}}$, not $c_{\text{in}}$. We price everything at its actual tier.
\end{itemize}

This reframes the question. Compression does not change $N$ directly and only shaves a slice off each $P_t$. So the interesting quantities are: \emph{how large a slice}, \emph{whether that slice grows with $t$}, and \emph{whether the intervention perturbs $N$}.

\section{Three Levers, Measured in Isolation}
\label{sec:levers}

\subsection{Lever 1 --- tool-schema filtering (the dominant, stable lever)}

Coding agents advertise their whole toolbox in full JSON schema on every request. Claude Code sends $\sim$31 tools ($\sim$29K tokens) with no MCP servers, and 60--90 tools (44--65K tokens) once MCP is attached. Most are irrelevant to the current step. \paritok{} keeps only the handful relevant to the agent's current intent in full schema and stubs the rest behind a recall tool, freezing the selection per conversation so the \texttt{tools[]} block stays byte-stable (cache-friendly).

The effect is large and reproducible. On single-file tasks with MCP attached, tool filtering saved \tok{392000} tokens over one \texttt{jinja2} session ($\sim$32K/request $\times$ 12) and \tok{1220000} tokens over a 34-request \texttt{rich} session ($\sim$36K/request), against content-compression savings of \tok{18000} and \tok{28000} respectively---a raw ratio of \textbf{20:1 to 33:1} in favor of tool filtering (Table~\ref{tab:ratio}).

\begin{table}[t]
\centering
\caption{Isolated per-session savings: tool filtering vs.\ content compression (MCP attached). Tool filtering is billed every request; content compression is counted once per unique file. Even after multiplying content compression by the turns it rides in history, tool filtering dominates.}
\label{tab:ratio}
\small
\begin{tabular}{lrrr}
\toprule
Session & Tool-filter saved & Content saved & Ratio \\
\midrule
\texttt{jinja2} (\texttt{filters.py}, 14K) & \tok{392000} & \tok{18000} & $\sim$20:1 \\
\texttt{rich} (\texttt{text.py}, 10K, 34 req) & \tok{1220000} & \tok{28000} & $\sim$33:1 \\
\bottomrule
\end{tabular}
\end{table}

A subtle bonus: some clients enlarge the tool block specifically when the endpoint is not the official host. Measured with Claude Code, \texttt{base\_url=api.anthropic.com} yields $\sim$40K tokens/turn of tools, while \texttt{base\_url=127.0.0.1} (any proxy) yields $\sim$57K/turn---a $\sim$17K/turn ``proxy tax.'' Tool filtering not only removes irrelevant tools but cancels this tax, which is why a filtered proxy can beat a direct connection outright ($\sim$\$0.043 vs.\ \$0.057 on a Sonnet edit task).

\textbf{Caveat for tool-poor agents.} Filtering must never stub an agent's core execution tool. Codex exposes only $\sim$9 tools and lives or dies by \texttt{shell\_command}; an embedding filter that intermittently drops it below top-$k$ paralyzes the agent (it stops running commands and asks the user to paste files). We whitelist a set of core execution tools (\texttt{shell}, \texttt{exec}, \texttt{apply\_patch}, \dots) that are never stubbed. With the whitelist, Codex runs commands every turn; without it, filtering is unsafe. Correspondingly, tool filtering saves Codex almost nothing (128--256 tokens): a 9-tool agent has no tool bloat to cut.

\subsection{Lever 2 --- content compression (a marginal per-turn effect)}

Compressing file reads to 26\% of their size sounds decisive but is, per turn, small: most of a turn's prefix is the fixed system+tools+history, and the compressible file slice is a few percent of it. In a turn-aligned comparison (GPU backend, deduplicated by message id), the compressed and uncompressed sides are \emph{identical} on turn~0 (before any file is read), and thereafter differ by only $\sim$1.5--2K tokens/turn---exactly the compressed file slice, entirely in the cheap cache-read tier (Table~\ref{tab:turnaligned}).

\begin{table}[t]
\centering
\caption{Turn-aligned prefix, content compression only (Sonnet/Claude Code, \texttt{sparkline} task). The per-turn saving is the compressed file slice ($\sim$2K), all cache-read.}
\label{tab:turnaligned}
\small
\begin{tabular}{lrrr}
\toprule
Turn & Compress input & No-op input & $\Delta$ \\
\midrule
0 & \tok{23275} & \tok{23275} & 0 \\
6 & \tok{34968} & \tok{36617} & $-$\tok{1649} \\
8 & \tok{36612} & \tok{38744} & $-$\tok{2132} \\
\bottomrule
\end{tabular}
\end{table}

Because this $\sim$2\%/turn effect is smaller than the turn-count variance, whole-task comparisons of ``compress vs.\ no-op'' come out \emph{time-varying in sign}: in one multi-run block the compressed side cost \$0.638 (18 turns) vs.\ \$0.230 (7 turns) uncompressed on the local backend, and the reverse on the GPU backend (\$0.268 at 10 turns vs.\ \$0.469 at 16). The difference is entirely turn count (7/10/13/16/18), not compression. \emph{Any} claim about content compression measured on a handful of runs is dominated by this variance; only distributions over $\ge 3$ runs are trustworthy.

\subsection{Lever 3 --- history summarization}

Summarizing stale turns keeps a long session inside the model's context window rather than overflowing it (or forcing an aggressive client-side compaction that drops detail). Its value is capacity, not primarily dollars, and it interacts with the scaling behavior we analyze next: by shrinking each turn's prefix, compression and summarization together let the agent fit \emph{more} turns before hitting the window---effectively buying back context length.

\section{Scaling Laws: Linear vs.\ Quadratic}
\label{sec:scaling}

The two content-bearing levers save at different rates as a session lengthens, and this is the key to when each one matters. We ran the \emph{same} read-only ``find the bug'' task for five consecutive turns in one Claude Code session (\texttt{-{}-resume}), with tool filtering left on for both arms so the A/B isolates content compression. Table~\ref{tab:cumulative} gives the cumulative savings.

\begin{table}[t]
\centering
\caption{Five consecutive read-only turns, one session, with tool filtering left on for \emph{both} arms so the A/B isolates content compression. ``Saved'' is the measured input difference (No-op $-$ Compress); its running sum fits $\approx 3{,}350\,N^2$. The tool-filter saving is not in this A/B (on for both arms); it is the fixed $\sim$21K/turn established in \S\ref{sec:levers}.}
\label{tab:cumulative}
\small
\begin{tabular}{crrrr}
\toprule
Turn $N$ & Compress in. & No-op in. & Saved (turn) & Saved (cum.) \\
\midrule
1 & \tok{72041} & \tok{75507} & \tok{3466}  & \tok{3466} \\
2 & \tok{51093} & \tok{61678} & \tok{10585} & \tok{14051} \\
3 & \tok{53877} & \tok{71538} & \tok{17661} & \tok{31712} \\
4 & \tok{56681} & \tok{81381} & \tok{24700} & \tok{56412} \\
5 & \tok{59697} & \tok{86995} & \tok{27298} & \tok{83710} \\
\midrule
\multicolumn{4}{r}{cumulative content saving at $N{=}5$:} & \tok{83710}\ ($22.2\%$)\\
\bottomrule
\end{tabular}
\end{table}

\paragraph{Content compression is quadratic.} The cumulative measured saving (No-op $-$ Compress input) is
\[
3466,\ 14051,\ 31712,\ 56412,\ 83710,
\]
which fits $c\,N^2$ with $c\approx 3{,}350$ (the per-$N^2$ ratios lie in $3.35$--$3.53{\times}10^3$; the turn-5 point gives $83710/25 = 3348$, reproducing the $22.2\%$ headline). The mechanism is compounding: each turn reads a file whose compressed form saves a few thousand tokens \emph{that turn}, and because the compressed block stays in history and is re-sent (cache-read) on every later turn, the per-turn saving itself grows roughly linearly ($\sim$7K/turn) while its running sum grows quadratically, $\approx 3{,}350\,N^2$.

\paragraph{Tool filtering is linear.} From \S\ref{sec:levers}, the unfiltered tool block is $\sim$29K tokens/turn and the filtered block $\sim$8K, so tool filtering removes a fixed $\sim$21K tokens \emph{every} turn, independent of conversation length; cumulative $\approx 21{,}000\,N$. (This lever is not visible in Table~\ref{tab:cumulative}, whose A/B keeps the filter on for both arms; it is the per-turn mechanism, measured against a direct no-proxy baseline.)

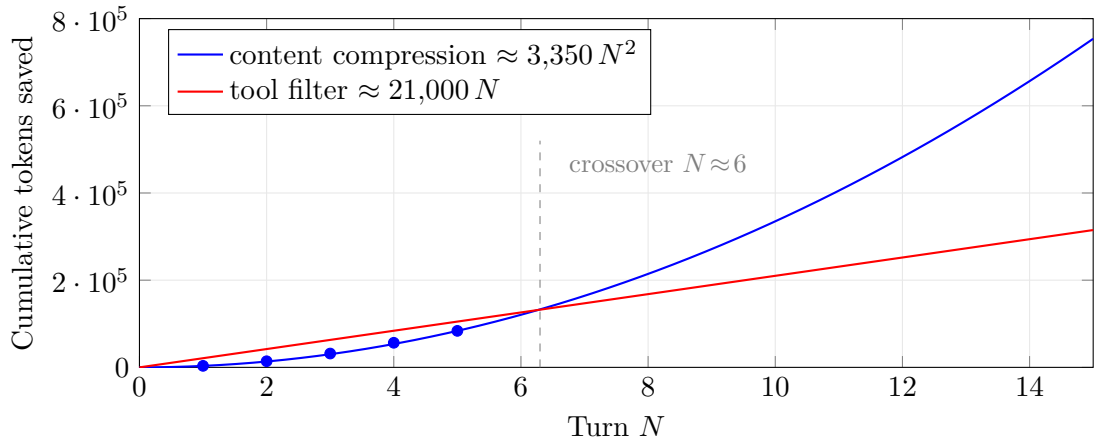
\begin{figure}[t]
\centering
\begin{tikzpicture}
\begin{axis}[
  width=0.86\linewidth, height=6.2cm,
  xlabel={Turn $N$}, ylabel={Cumulative tokens saved},
  xmin=0, xmax=15, ymin=0, ymax=800000,
  legend pos=north west, legend cell align=left,
  grid=both, grid style={gray!18},
  scaled y ticks=false, yticklabel style={/pgf/number format/1000 sep={,}},
]
\addplot[thick, blue, domain=0:15, samples=60] {3350*x^2};
\addlegendentry{content compression $\approx 3{,}350\,N^2$}
\addplot[thick, red, domain=0:15, samples=2] {21000*x};
\addlegendentry{tool filter $\approx 21{,}000\,N$}
\addplot[only marks, mark=*, blue] coordinates {(1,3466)(2,14051)(3,31712)(4,56412)(5,83710)};
\draw[dashed, gray] (axis cs:6.3,0) -- (axis cs:6.3,520000);
\node[gray, anchor=west] at (axis cs:6.6,470000) {\small crossover $N\!\approx\!6$};
\end{axis}
\end{tikzpicture}
\caption{Measured content-compression savings (markers, from the Table~\ref{tab:cumulative} A/B) and fitted scaling laws (curves). The tool filter is a fixed per-turn cut (linear); content compression compounds (quadratic) and overtakes near turn~6, before the $\sim$200K context window caps the quadratic around turn 12--15.}
\label{fig:scaling}
\end{figure}

\paragraph{Crossover and cap.} Setting $3{,}350\,N^2 = 21{,}000\,N$ gives $N\approx 6$: the fixed tool-filter saving dominates only for the first $\sim$6 turns; past that, the compounding content saving overtakes it and the gap widens quadratically (Figure~\ref{fig:scaling}). But the quadratic does not run forever. The uncompressed prefix is already 87K/turn at turn~5 and grows $\sim$10--15K/turn, so an unassisted session saturates the $\sim$200K window around turn 12--15, at which point client-side compaction truncates history and the accumulation flattens---we already see the per-turn delta's growth slow at turn~5 ($+2598$ vs.\ $\sim$$7000$ earlier). Thus: content compression's saving grows quadratically \emph{until the window caps it around turn 12--15}, and its own effect (smaller prefixes) pushes that cap outward.

This is the paper's central quantitative result: \emph{the two levers are not substitutes but complements with different growth exponents}. On a single turn the tool filter is the only thing worth measuring; on a long session where the agent seldom recalls, content compression eventually becomes the larger lever.

\section{The Recall Trap: What Recall Costs Content Compression}
\label{sec:expand}

A non-destructive gateway's defining feature is that nothing is lost: the agent can recall exact original bytes on demand. Recall is \emph{momentary}: the gateway surfaces the original only for the turn that asks for it and re-stubs it to its \texttt{[REF]} on the very next turn---the agent reads the exact bytes, acts on them, and the original is immediately dropped from the context again (it never lingers to re-inflate later prefixes). This is essential for correctness---but it is also precisely what makes each recall cost a fixed, bounded amount---one compressed segment---whenever the agent recalls the original to obtain those exact bytes.

\subsection{Isolated evidence: recall costs one compressed segment}

We isolated content compression completely (MCP off, tool filter off---so only the file-compression path differs) on a code-change task (\texttt{sparkline}). Content compression shrinks each read by only $\sim$2\% of the cache-priced prefix; but to apply a precise change the agent needs the \emph{exact} bytes, so it recalls the original---and a recall simply re-sends the segment that was just compressed. The moment the original returns, the compressed copy the gateway had stored becomes pure waste, of \emph{fixed} size equal to the length of that compressed text. This is not an error effect: with the edit-recovery boundary bug fixed (below) there were zero reflow typos and \texttt{Edit} exact-match failures were $0$ across all runs, so the recalls are how the agent obtains exact bytes, not a symptom of broken edits. The cost of recall is therefore bounded and deterministic---one wasted compressed segment per recall---so wherever the agent recalls, the price is exactly that one compressed segment, not a multiplicative cost increase.

\subsection{The recall economics}

Each recall wastes a \emph{fixed}, bounded amount---the compressed copy of the one segment whose original it re-sends, since compressing that segment then bought nothing. It is \emph{not} a fresh full-prefix round-trip: the accumulated $\sim$35K prefix is already cached (cache-read, $\sim$0.1$\times$) and would be re-sent next turn regardless, so the only new cost a recall adds is that single compressed segment. Crucially the recalled original is \emph{ephemeral}: the gateway resolves it only for the turn that asks and re-stubs it to its \texttt{[REF]} on the next turn, so it never persists to re-inflate the prefix---the segment does \emph{not} revert to full size for the rest of the session, and the recall stays a one-time cost. A recall therefore costs exactly one compressed segment---not a blown-up bill---and the content lever's net is the accumulated saving minus one such segment per recall. How often an agent recalls depends only on how much of the original the change needs:

\begin{itemize}[leftmargin=1.4em,itemsep=2pt]
  \item \textbf{Changes the summary already supports} (e.g.\ set \texttt{max\_retries=-1}): the exact value survives compression, the agent applies it directly, recalls${}=0$, and compression stays marginally positive ($\sim$6.5\% fewer total input tokens, all cache-read---dollar-neutral).
  \item \textbf{Changes that need the untouched bytes} (e.g.\ a precise multi-line replacement, where the 4B model had reflowed a retained signature onto one line): the agent recalls the original (stably 2 recalls), each recall costing one compressed segment---a small, bounded loss, not a blowup.
\end{itemize}

Recall must not be removed, though: deleting the recall tool entirely is far worse---the agent then reconstructs exact bytes via \texttt{grep}/\texttt{cat}/re-reads, reaching 46 turns / \$1.85, $3.2\times$ the cost of keeping recall.

\subsection{When content compression nets positive}

Combining \S\ref{sec:scaling} and \S\ref{sec:expand}: content compression's net is the quadratic accumulation \emph{minus} a bounded cost of one compressed segment per recall, so its sign is set by exactly two variables---\emph{session length} and \emph{how often the agent recalls the original}. It is positive when both favor it: the session is long (so the accumulation runs) \emph{and} recalls are rare (so little of the saving is spent back). What the task is called does not decide this---only the recall count does: a change the summary already supports needs no recall and stays positive, while one that pulls the original back costs a single compressed segment. The practical rule is \emph{use more, save more}, with the fine print \emph{many turns, few recalls}.

\section{A Benchmark Caveat: Single-Shot Quality $\perp$ Multi-Turn Cost}
\label{sec:swebench}

The compression \emph{model} the gateway deploys, Paritok-4B (reported separately), is strong on the standard benchmark: on SWE-bench Lite~\cite{swebench} it retains \textbf{86.5\%} of uncompressed solve quality at a \textbf{25.7\%} compression rate, matching a gpt-4.1-mini compressor (85.6\% at 50.2\%) at less than half the tokens (Table~\ref{tab:swe}). We reproduce that number here only to make a point about how it must \emph{not} be read: it is tempting to cite such a benchmark as evidence that ``compression saves tokens without hurting the agent.'' \emph{This inference is invalid,} and naming the reason is one of our contributions.

\begin{table}[t]
\centering
\caption{SWE-bench Lite, single-shot: context passed through each compressor, one API call, model emits a unified diff. Quality retained $=$ the compressed arm's solve quality normalized to the uncompressed baseline, both under the identical single-shot harness.}
\label{tab:swe}
\small
\begin{tabular}{lcc}
\toprule
Context source & Quality retained & Compression rate \\
\midrule
Uncompressed baseline & 100.0\% & 100.0\% \\
gpt-4.1-mini (compressor) & 85.6\% & 50.2\% \\
gpt-5 (compressor) & 93.6\% & 61.9\% \\
\textbf{Paritok-4B} & \textbf{86.5\%} & \textbf{25.7\%} \\
\bottomrule
\end{tabular}
\end{table}

The benchmark harness makes \emph{one} API call: it stuffs the (compressed or full) context plus the issue into a single user message and asks the model to emit a diff, applied with a fuzzy patcher. There are no tools, no turns, no re-reads, no exact-match \texttt{Edit}. It therefore measures \emph{single-shot understanding under compression}---and shows understanding survives heavy compression. But a real agent's cost gap is not in understanding; it is in \emph{multi-turn behavior and the tool block}, both of which the benchmark omits entirely: it carries no 44--65K MCP tool block (so it cannot see that tool filtering is the real lever), it never re-sends an accumulating prefix, and its fuzzy diff application hides the exact-match brittleness a real \texttt{Edit} tool exposes. Single-shot quality and multi-turn cost are orthogonal axes; a compressor can be excellent on the first and irrelevant---or harmful---on the second. The 86.5\% (reported separately for Paritok-4B) is best read as a \emph{floor} on understanding, not as a cost claim about the gateway.

\section{Methodology for Reproducible Agent-Cost Measurement}
\label{sec:method}

Our strongest finding about method is that \emph{per-turn attribution is unreliable}; agent cost is variance-dominated and only distributions are trustworthy. Concretely:

\begin{itemize}[leftmargin=1.4em,itemsep=2pt]
  \item \textbf{Report distributions, not single runs.} The same task swings 2$\times$ in turn count and cost. We run $\ge 3$ trials per arm and compare distributions; conclusions drawn from one run repeatedly reversed themselves.
  \item \textbf{Use provider-reported \texttt{usage}, priced per tier.} Cache-read is $\sim$0.1$\times$ the fresh-input price; a saving's tier matters as much as its size.
  \item \textbf{Control the tool block.} MCP servers load asynchronously and non-deterministically (we observed 40 vs.\ 90 tools across two concurrent runs, a 40K first-turn swing). This alone can masquerade as a compression effect; we pin it with \texttt{-{}-strict-mcp-config} when isolating other levers.
  \item \textbf{Deduplicate streamed usage.} Client transcripts contain duplicated streaming entries; aggregate by message id. Server-side recall is invisible in the client transcript and must be counted via the gateway's own logs.
  \item \textbf{Isolate one lever at a time} via the \texttt{min\_tokens} floor (content), a filter toggle (tools), and \texttt{strict-mcp} (block size)---otherwise the proxy tax and MCP jitter contaminate the comparison.
\end{itemize}

We also note a genuine code-level pitfall surfaced by this study: an edit-recovery routine that re-aligns a reflowed \texttt{old\_string} to the real multi-line file had an asymmetric boundary rule (prefix reclaimed only \texttt{"\ \textbackslash t"}, suffix used \texttt{.isspace()}), gluing two tokens across a newline (\texttt{max\_widthheight}). Making both boundaries use \texttt{.isspace()} fixed it (24 recovery tests pass). This is orthogonal to the cost findings but illustrates that non-destructive recovery has its own correctness surface.

\section{Recommendations}
\label{sec:recs}

For practitioners deciding where to spend token-saving effort on a multi-turn coding agent:

\begin{enumerate}[leftmargin=1.4em,itemsep=2pt]
  \item \textbf{Filter the tool schema first.} It is the largest per-turn lever, the only reliably-positive one, and it grows the agent's toolbox (MCP) without growing the bill. Keep the selection frozen per conversation for cache stability, and never stub the agent's core execution tool.
  \item \textbf{Treat content compression as a session-length bet, not a per-turn win.} It pays off super-linearly on long sessions where the agent seldom recalls the original (e.g.\ auditing or Q\&A over a large codebase), and on short sessions or the minority of turns where the agent recalls it costs at most one compressed segment---bounded, never a multiplicative loss.
  \item \textbf{Keep compression non-destructive, but budget for recall.} Exact recovery is required for correctness; just subtract one compressed segment's saving on any turn where the agent recalls.
  \item \textbf{Report end-to-end dollars at the correct cache tier}, and never cite a single-shot benchmark as a multi-turn cost result.
  \item \textbf{The next frontier for file-level value is a symbol map / directed retrieval}---be the agent's \texttt{grep} (which functions exist, at which lines, fetch the exact span on demand)---rather than indiscriminate body deletion, which agents route around by slice-reading and which harms editing.
\end{enumerate}

\section{Related Work and Limitations}

Prompt-compression methods---token pruning (selective-context~\cite{selectivecontext}, the LLMLingua family~\cite{llmlingua,llmlingua2}) and soft-prompt compression (gist tokens~\cite{gist}, AutoCompressors~\cite{autocompressors})---optimize single-prompt compression rate and downstream quality, the same axis as our SWE-bench measurement; our contribution is to show that this axis is orthogonal to multi-turn agent \emph{cost}. Gateway/proxy context managers that summarize conversation history address lever~3 but leave the file/tool content---the fastest-growing part of the bill---untouched; skeleton-extraction tools reduce reads to signatures but force the re-expansion round-trips our recall analysis quantifies.

Our study has clear limits. Numbers are from a specific 4B compressor, two agent families, and Python-heavy repositories; the scaling \emph{exponents} (linear/quadratic) should generalize, but constants will not. The quadratic fit rests on five turns; while the mechanism (re-sent accumulating history) makes the $N^2$ form near-inevitable, the cap's exact turn depends on window size and client compaction policy. Codex-on-Windows introduced shell-encoding noise orthogonal to compression. We report ranges and distributions accordingly.

\section{Conclusion}

``Compress the file reads'' is the intuitive way to cut a coding agent's token bill, and it is mostly the wrong lever. We decomposed real multi-turn agent cost and found that (i) turn count, not compression rate, sets the bill; (ii) tool-schema filtering is the dominant, reproducibly-positive lever, saving a fixed block linearly every turn; (iii) content compression is a $\sim$2\% per-turn effect that nonetheless accumulates \emph{quadratically} over a long session and overtakes the tool filter near turn~6 before the context window caps it; (iv) each recall costs a bounded, fixed amount---one compressed segment---eroding the content saving only in proportion to the (minority) recall count; and (v) single-shot compression benchmarks are orthogonal to multi-turn cost. The practical takeaway is a hierarchy: filter tools always, compress content when the session is long and the agent rarely recalls, and measure in end-to-end dollars at the right cache tier. Decomposition, not a headline compression rate, is what tells you where the tokens actually go.

\paragraph{Reproducibility.} The gateway, the 4B compression model, and the evaluation harness are open source (Apache~2.0) at \url{https://github.com/Paritok-official/paritok-4b-v1}; the model weights are on the Hugging Face Hub.

\end{document}